\documentclass[letterpaper, 10 pt, conference]{ieeeconf}

\IEEEoverridecommandlockouts                              %
\usepackage[T1]{fontenc}
\usepackage[utf8]{inputenc}

\usepackage{amsmath} %
\usepackage{amssymb} %

\usepackage[hidelinks]{hyperref}
\usepackage{graphicx}
\usepackage{tabularx}
\usepackage[table]{xcolor}
\definecolor{mygreen}{HTML}{62A260}
\definecolor{myyellow}{HTML}{FAF7C2}
\definecolor{myred}{HTML}{FFA194}
\definecolor{todored}{HTML}{C0392B}

\usepackage{tikz}
\usetikzlibrary{positioning,fit,backgrounds,calc,arrows.meta}

\usepackage[
    frozencache,
    cachedir=minted-cache
]{minted}
\setminted{
  linenos,
  breaklines=true,
  fontsize=\scriptsize,
  numbersep=4pt,
  xleftmargin=1.4em,
  framesep=2pt
}
\renewcommand{\listingscaption}{Code snippet}

\makeatletter
\let\@float@c@listing\@caption
\makeatother

\usepackage{multirow}
\usepackage{booktabs}

\usepackage{cleveref}

\usepackage{bm}
\usepackage{url}

\usepackage{ifthen}

\usepackage{xspace}
\newcommand{\ecv}{\mbox{EventCV}\xspace}
\newcommand{\code}[1]{\texttt{\small #1}}

\newif\ifpendingresults
\pendingresultstrue

\newif\ifanonymous
\anonymousfalse

\title{\LARGE \bf The \ecv Library for Event-Based Robotic Vision}

\ifanonymous
  \author{Anonymous Author(s)%
  }
\else
  \author{Adam D. Hines$^{*}$ \quad Michael Milford \quad Tobias Fischer%
  \thanks{The authors are with the QUT Centre for Robotics, School of Electrical
  Engineering and Robotics, Queensland University of Technology, Brisbane, QLD 4000,
  Australia. $^{*}$Corresponding author: \texttt{adam.hines@qut.edu.au}}%
  \thanks{This work received funding from an ARC Laureate Fellowship FL210100156 to MM and an ARC Discovery Early Career Researcher Award DE240100149 to TF. The authors acknowledge continued support from the Queensland University of Technology (QUT) through the Centre for Robotics.}
  }
\fi

\begin{document}
\bstctlcite{IEEEexample:BSTcontrol}
\maketitle
\thispagestyle{empty}
\pagestyle{empty}

\begin{abstract}%
\label{sec:abstract}%
Event cameras detect per-pixel brightness changes asynchronously on microsecond timescales, with high dynamic range and low power draw. These are desirable properties for robots that move fast or work in difficult lighting conditions. However, integrating an event camera into a real-world robotic pipeline still requires substantial effort: plug-and-play drivers do not exist, event streams are recorded in a variety of incompatible file formats, and most projects rely on custom research-grade code. Here, we present \emph{\ecv}, an open-source and extensible Rust library with OpenCV-style Python bindings that lowers the entry barrier to working with event cameras. \ecv provides a wide range of features: denoising filters and geometric transforms, augmentations, corner detection and unsupervised feature learning, contrast-maximization motion estimation, a video-to-events simulator, and Open Neural Network Exchange (ONNX) inference for deployment in robotic stacks. \ecv integrates the Neuromorphic Drivers package, allowing an event camera stream to be processed directly in real time. No existing toolkit covers this range of operations in one package, and \ecv builds representations and decodes files 1.1$\times$ to 3.7$\times$ faster than the currently available libraries. We deploy \ecv on a Jetson Orin AGX and present three robotics case studies spanning object detection, on-device model inference, and localization. Project webpage: \url{https://eventcv.net}.
\end{abstract}

\section{Introduction}
\label{sec:intro}

Visual sensing in robotics is dominated by frame-based RGB cameras, which have decades of accumulated software development. A roboticist who buys a webcam plugs it in, calls \code{cv2.VideoCapture} from OpenCV~\cite{Bradski2000}, and is a single function call away from using images with popular computing libraries such as NumPy~\cite{Harris2020}, PyTorch~\cite{Paszke2019}, or the Machine Vision Toolbox~\cite{Corke2005}. Swapping cameras rarely alters the workflow: file formats are standardized and conversion into machine learning pipelines is routine.

An event camera, by contrast, records independent changes in log intensity at the pixel level and emits an event \mbox{$e=(x,y,t,p)$} for each such change, consisting of pixel coordinates $(x,y)$, a timestamp $t$, and a polarity $p$ that indicates whether there was a brightness increase or decrease~\cite{Lichtsteiner2008,Brandli2014}. The sensor resolves these events to microseconds, has a dynamic range beyond 120\,dB, and produces a data rate that follows scene activity instead of a fixed frame rate. Those properties suit robots that move fast or face abrupt changes in lighting, and the past decade has produced a large body of work including event-based odometry, optical flow, object recognition, and localization~\cite{Gallego2022}.

\begin{figure}[t]
    \centering
    \includegraphics[width=\columnwidth]{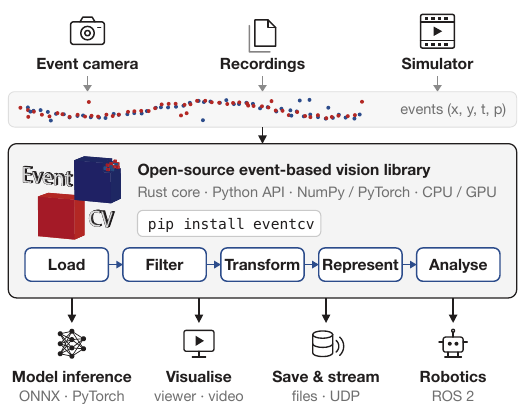}
    \caption{The \ecv ecosystem provides efficient camera and file streaming and integrates into modern machine learning and robotics workflows.}
    \label{fig:hero}
\end{figure}

Deploying event cameras on robots does not currently benefit from the same level of accessibility as conventional cameras, with several key challenges limiting the utility of available libraries. Existing event-based vision libraries typically address only part of the processing pipeline. In particular, common robotics operations such as geometric transformations, feature detection, motion estimation, and feature tracking are inconsistently supported, often requiring researchers to combine multiple libraries or custom implementations. There is also a divide between offline experimentation and live deployment, where processing pipelines developed for recorded datasets cannot necessarily be transferred directly to a live camera using the same interface. Finally, event cameras are not plug-and-play, as drivers do not self-install over USB, with the two largest vendors providing separate and incompatible software development kits (SDKs)~\cite{dvprocessing,metavision}.

    \begin{table*}[!t]
        \centering
        \caption{Comparison of event-based vision libraries and open-source projects. One, two, and three check marks denote basic, substantial, and comprehensive support for a capability. A cross denotes that it is absent.}
        \label{tab:comparison}
        \resizebox{\textwidth}{!}{%

        \begin{tabular}{@{}l@{\hspace{6pt}}c@{\hspace{6pt}}c@{\hspace{6pt}}c@{\hspace{6pt}}c@{\hspace{6pt}}c@{\hspace{6pt}}c@{\hspace{6pt}}c@{\hspace{6pt}}c@{\hspace{6pt}}c@{}}

        \toprule
        \textbf{Capability} & \textbf{EventCV} & \textbf{AEDat}~\cite{aedat} & \textbf{AEStream}~\cite{Pedersen2023} & \textbf{evlib}~\cite{evlib} & \textbf{Event utils}~\cite{Stoffregen2019} & \textbf{Expelliarmus}~\cite{expelliarmus} & \textbf{Faery}~\cite{faery} & \textbf{jAER}~\cite{Delbruck2008} & \textbf{Tonic}~\cite{Lenz2021} \\

        \midrule

        \textbf{Read/write}
        & \cellcolor{mygreen!40}{\large\checkmark\checkmark\checkmark}
        & \cellcolor{myyellow!40}{\large\checkmark}
        & \cellcolor{myyellow!40}{\large\checkmark\checkmark}
        & \cellcolor{myyellow!40}{\large\checkmark}
        & \cellcolor{myyellow!40}{\large\checkmark}
        & \cellcolor{myyellow!40}{\large\checkmark}
        & \cellcolor{myyellow!40}{\large\checkmark\checkmark}
        & \cellcolor{myyellow!40}{\large\checkmark\checkmark}
        & \cellcolor{myyellow!40}{\large\checkmark}\\
        \addlinespace[2pt]

        \textit{Format scope}
        & \textit{Broad I/O}
        & \textit{AEDAT4 read}
        & \textit{Selected formats}
        & \textit{Selected formats}
        & \textit{Research formats}
        & \textit{Prophesee formats}
        & \textit{Several formats}
        & \textit{Several formats}
        & \textit{Dataset parsers}  \\

        \textbf{Live streaming}
        & \cellcolor{mygreen!40}{\large\checkmark\checkmark\checkmark}
        & \cellcolor{myred!40}{\large\texttimes}
        & \cellcolor{myyellow!40}{\large\checkmark\checkmark}
        & \cellcolor{myred!40}{\large\texttimes}
        & \cellcolor{myred!40}{\large\texttimes}
        & \cellcolor{myred!40}{\large\texttimes}
        & \cellcolor{myyellow!40}{\large\checkmark\checkmark}
        & \cellcolor{mygreen!40}{\large\checkmark\checkmark\checkmark}
        & \cellcolor{myred!40}{\large\texttimes} \\
        \addlinespace[2pt]

        \textit{Sensor control}
        & \textit{Bias + ROI}
        & \textit{Not applicable}
        & \textit{Driver dependent}
        & \textit{Not applicable}
        & \textit{Not applicable}
        & \textit{Not applicable}
        & \textit{Driver dependent}
        & \textit{Bias controls}
        & \textit{Not applicable} \\

        \textbf{Filter/transforms}
        & \cellcolor{mygreen!40}{\large\checkmark\checkmark\checkmark}
        & \cellcolor{myred!40}{\large\texttimes}
        & \cellcolor{myyellow!40}{\large\checkmark}
        & \cellcolor{myyellow!40}{\large\checkmark}
        & \cellcolor{myyellow!40}{\large\checkmark\checkmark}
        & \cellcolor{myyellow!40}{\large\checkmark}
        & \cellcolor{myyellow!40}{\large\checkmark}
        & \cellcolor{mygreen!40}{\large\checkmark\checkmark\checkmark}
        & \cellcolor{myyellow!40}{\large\checkmark\checkmark} \\
        \addlinespace[2pt]

        \textit{Event processing}
        & \textit{Geometry + filters}
        & \textit{Not provided}
        & \textit{Basic transforms}
        & \textit{Lazy filtering}
        & \textit{Augment + motion}
        & \textit{Windowed I/O}
        & \textit{Stream operations}
        & \textit{Broad filter stack}
        & \textit{Augmentation suite} \\

        \textbf{Representations}
        & \cellcolor{mygreen!40}{\large\checkmark\checkmark\checkmark}
        & \cellcolor{myred!40}{\large\texttimes}
        & \cellcolor{myyellow!40}{\large\checkmark}
        & \cellcolor{mygreen!40}{\large\checkmark\checkmark\checkmark}
        & \cellcolor{myyellow!40}{\large\checkmark\checkmark}
        & \cellcolor{myred!40}{\large\texttimes}
        & \cellcolor{myyellow!40}{\large\checkmark}
        & \cellcolor{myyellow!40}{\large\checkmark}
        & \cellcolor{myyellow!40}{\large\checkmark\checkmark} \\
        \addlinespace[2pt]

        \textit{Representation scope}
        & \textit{Multiple families}
        & \textit{Not provided}
        & \textit{Dense tensors}
        & \textit{Multiple families}
        & \textit{Core families}
        & \textit{Not provided}
        & \textit{Rendered frames}
        & \textit{Filter outputs}
        & \textit{Core families} \\

        \textbf{Feature tracking}
        & \cellcolor{mygreen!40}{\large\checkmark\checkmark\checkmark}
        & \cellcolor{myred!40}{\large\texttimes}
        & \cellcolor{myred!40}{\large\texttimes}
        & \cellcolor{myred!40}{\large\texttimes}
        & \cellcolor{myyellow!40}{\large\checkmark}
        & \cellcolor{myred!40}{\large\texttimes}
        & \cellcolor{myred!40}{\large\texttimes}
        & \cellcolor{mygreen!40}{\large\checkmark\checkmark\checkmark}
        & \cellcolor{myred!40}{\large\texttimes} \\
        \addlinespace[2pt]

        \textit{Tracking scope}
        & \textit{Corners + features}
        & \textit{Not provided}
        & \textit{Not provided}
        & \textit{Not provided}
        & \textit{Motion segmentation}
        & \textit{Not provided}
        & \textit{Not provided}
        & \textit{Tracking filters}
        & \textit{Not provided} \\

        \textbf{Event simulation}
        & \cellcolor{mygreen!40}{\large\checkmark\checkmark\checkmark}
        & \cellcolor{myyellow!40}{\large\checkmark}
        & \cellcolor{myred!40}{\large\texttimes}
        & \cellcolor{myyellow!40}{\large\checkmark\checkmark}
        & \cellcolor{myyellow!40}{\large\checkmark}
        & \cellcolor{myred!40}{\large\texttimes}
        & \cellcolor{myyellow!40}{\large\checkmark}
        & \cellcolor{myyellow!40}{\large\checkmark}
        & \cellcolor{myyellow!40}{\large\checkmark} \\
        \addlinespace[2pt]

        \textit{Simulation scope}
        & \textit{Sensor model}
        & \textit{Not provided}
        & \textit{Not provided}
        & \textit{ESIM~\cite{Rebecq2018esim} model}
        & \textit{Noise synthesis}
        & \textit{Not provided}
        & \textit{Synthetic input}
        & \textit{External v2e}
        & \textit{Noise injection} \\

        \textbf{Visualization/analytics}
        & \cellcolor{mygreen!40}{\large\checkmark\checkmark\checkmark}
        & \cellcolor{myred!40}{\large\texttimes}
        & \cellcolor{myyellow!40}{\large\checkmark\checkmark}
        & \cellcolor{myyellow!40}{\large\checkmark}
        & \cellcolor{myyellow!40}{\large\checkmark\checkmark}
        & \cellcolor{myred!40}{\large\texttimes}
        & \cellcolor{mygreen!40}{\large\checkmark\checkmark\checkmark}
        & \cellcolor{mygreen!40}{\large\checkmark\checkmark\checkmark}
        & \cellcolor{mygreen!40}{\large\checkmark\checkmark\checkmark} \\
        \addlinespace[2pt]

        \textit{Analysis scope}
        & \textit{Viewer + analytics}
        & \textit{External tools}
        & \textit{Live viewer}
        & \textit{Figure scripts}
        & \textit{Plots + video}
        & \textit{Not provided}
        & \textit{Rendered video}
        & \textit{Desktop analysis}
        & \textit{Plots + animation} \\

        \textbf{Compute scale}
        & \cellcolor{mygreen!40}{\large\checkmark\checkmark\checkmark}
        & \cellcolor{myred!40}{\large\texttimes}
        & \cellcolor{mygreen!40}{\large\checkmark\checkmark\checkmark}
        & \cellcolor{mygreen!40}{\large\checkmark\checkmark\checkmark}
        & \cellcolor{myyellow!40}{\large\checkmark\checkmark}
        & \cellcolor{myyellow!40}{\large\checkmark}
        & \cellcolor{myyellow!40}{\large\checkmark\checkmark}
        & \cellcolor{mygreen!40}{\large\checkmark\checkmark\checkmark}
        & \cellcolor{myyellow!40}{\large\checkmark} \\
        \addlinespace[2pt]

        \textit{Compute path}
        & \textit{Rust + GPU}
        & \textit{Decoder only}
        & \textit{C++ + CUDA}
        & \textit{Rust + GPU}
        & \textit{NumPy + PyTorch}
        & \textit{Chunked decoding}
        & \textit{Streaming pipeline}
        & \textit{Java real-time}
        & \textit{Python + NumPy} \\

        \bottomrule
        \end{tabular}
        }
    \footnotesize
    \end{table*}

In this work, we introduce \emph{\ecv} (Fig.~\ref{fig:hero}), an open-source computer vision library for event cameras. \ecv uses a computationally efficient Rust core with an OpenCV-style Python frontend for use in modern machine learning workflows. Existing libraries (Table~\ref{tab:comparison}) have laid the groundwork for resolving individual challenges in file format decoding, camera driver development, and event stream processing. \ecv unifies many of these features into a single Python package, reducing the dependency on multiple libraries. jAER~\cite{Delbruck2008} is the closest related work to \ecv in terms of breadth and scope, with the main architectural distinction being its Java-based implementation compared with \ecv's Rust backend and Python interface. \ecv couples to the Neuromorphic Drivers package~\cite{neuromorphicdrivers}, so installing it gives immediate access to a supported camera. \ecv is designed so that simple augmentations, transforms, and representations are a single function call.

Specifically, we contribute:

\begin{enumerate}
    \item The \ecv Python library\footnote{Open-source code at \url{https://github.com/EventLAB-Team/eventcv}}, providing camera interfacing, I/O across multiple formats, filtering, transforms, representations, feature detection, motion estimation, simulation, and model inference.
    \item A timing evaluation of \ecv's representation building and file decoding against existing libraries, with the Rust backend 1.1$\times$ to 3.7$\times$ faster than the next-fastest library.
    \item Three robotics case studies that rebuild existing workflows with \ecv, spanning object detection, real-time model inference, and a standardized localization pipeline.
\end{enumerate}

\section{Related Work}
\label{sec:relatedworks}

To date, there are several open-source tools that partially cover the breadth of event camera data handling, and we provide an overview of them here. Table~\ref{tab:comparison} lists the most common libraries available and rates their coverage of capabilities.

\subsection{Event stream processing}
\label{subsec:streamproc}
Commercial SDKs provide comprehensive event camera streaming and processing capabilities, but are limited to their own sensor family. iniVation's \emph{dv-processing}~\cite{dvprocessing} offers capture, noise filtering, accumulators, time surfaces, and feature tracking for the DAVIS and DVXplorer lines of cameras. Prophesee's \emph{Metavision SDK}~\cite{metavision}, whose open-source subset was OpenEB until it was archived, does the same for Prophesee sensors and adds a video-to-events simulator.

\emph{AEStream}~\cite{Pedersen2023} is a high-throughput system that moves events from a camera or file to a GPU, file, or network endpoint. A critical component of AEStream is a User Datagram Protocol (UDP) transport that allows connection to SpiNNaker boards. \emph{Faery}~\cite{faery} moves events between cameras, files, and network endpoints, and renders representations from them. \emph{jAER}~\cite{Delbruck2008} is the longest-standing open-source library with over two decades of real-time streaming functionality and filters. However, jAER is built on a Java platform that modern Python stacks cannot import, and Java is no longer widely used for scientific computing. At the same time, jAER remains considerably more mature at the hardware level, providing broader sensor support together with extensive bias configuration, device-specific controls, and real-time sensor-processing tools. jAER also provides real-time feature extraction, tracking, optical flow, denoising, sensory-motor processing, and robot demonstrators. \emph{evlib}~\cite{evlib} is the most similar to \ecv in construction, i.e., a Rust core with Python bindings for event stream handling. It uses lazy evaluation over the Python Polars package, dense representations, PyTorch datasets, and an ESIM-based~\cite{Rebecq2018esim} simulator.

For robotics applications, the \emph{event-driven} library for the Yet Another Robot Platform (YARP) middleware~\cite{Glover2018yarp} is the most directly comparable middleware integration. In ROS, \emph{rpg\_dvs\_ros}~\cite{Mueggler2017ros} supplies the DVS/DAVIS driver and the \code{dvs\_msgs} bag layout. A maintained ROS~2 stack~\cite{rosevent} adds a vendor-neutral message format, codecs, and drivers.

\subsection{Dataset and file handling}
\label{subsec:datafile}
Event recordings are distributed in file formats that cannot be read interchangeably. For example, AEDAT~2.0 and~4.0, Prophesee DAT and EVT2/EVT3 RAW, ROS bags of \code{dvs\_msgs}, HDF5, and plain text are among the most common file formats encountered for event streams. Some open-source tools handle individual formats: \emph{Expelliarmus}~\cite{expelliarmus} decodes Prophesee DAT and RAW, whereas the \emph{aedat} crate~\cite{aedat} decodes AEDAT~4.0 only. Tonic~\cite{Lenz2021} is a comprehensive library that handles downloading, caching, and versioning for roughly twenty standard datasets, including N-ImageNet dataset~\cite{Kim2021}.

\subsection{Feature detection and tracking}
\label{subsec:featdetect}
Several event-based feature extraction methods do not depend on deep-learned models and can be re-implemented in pure Rust. eFAST~\cite{Mueggler2017} and Arc~\cite{Alzugaray2018} detect corners on the surface of active events (SAE). FEAST~\cite{Afshar2020} learns feature prototypes online. Motion is most often recovered by contrast maximization, introduced for angular velocity estimation~\cite{Gallego2017} and generalized into a unifying framework~\cite{Gallego2018} whose reward functions are analysed in~\cite{Stoffregen2019,Stoffregen2021}. Several learned methods estimate flow and egomotion directly~\cite{Zhu2019}.

\ecv differs from these tools in combining them: one package that streams from a camera, reads nine file formats, and provides filters, representations, features, motion estimation, simulation, and model inference behind a single interface.

\section{Design and Architecture}
\label{sec:archcompat}

\subsection{Overview}
\label{subsec:overview}

\ecv uses a computationally efficient Rust core, \code{eventcv-core}, that is connected to a thin Python API layer. The Rust core handles the expensive parts of event stream processing and parallelizes them. The components include slice indexing and decoding, camera streaming threads, representation building, and filter and estimator kernels. Event data can easily be converted into conventional array structures through the Python API by calling \code{.numpy()} on \ecv Rust objects. Inspired by the functionality of libraries such as OpenCV, \ecv offers both functions and bound methods that are chainable by design. For example, generating a voxel representation from an event stream is callable either as \code{ecv.voxel(stream, bins=5)} or \code{stream.voxel(bins=5)}, which are functionally equivalent.

To easily access event streams from a physically connected event camera, \ecv depends on the Neuromorphic Drivers package~\cite{neuromorphicdrivers}, which provides the ability to directly stream and record from cameras such as the Prophesee EVK3~HD and EVK4, the iniVation DVXplorer and DAVIS346, the SilkyEvCam~HD, and the IDS uEye XCP-E. Similarly, ROS~2 transport is enabled via \emph{hiroz} (Sec.~\ref{subsec:ros2}), and deep learning model execution can be achieved via integration with the Open Neural Network Exchange (ONNX) Runtime. \ecv itself implements the decoding, slicing, filtering, representation, feature, and motion code around these libraries.

Installing \ecv into a Python environment provides direct access to streaming from an event camera without any additional setup (Code snippet~\ref{lst:eventcv_stream}). This snippet shows the complete setup required to plug in an event camera via USB and begin streaming. The alternative is vendor SDKs tied to one sensor family, or custom code.

\begin{listing}[t]
\begin{minted}[frame=lines, breaklines, breakanywhere]{console}
# Install Pixi
curl -fsSL https://pixi.sh/install.sh | sh
# Create a new project and add eventcv as dependency
pixi init && pixi add eventcv
pixi run python
# Run Python, import EventCV, and start streaming
import eventcv as ecv
ecv.stream()
\end{minted}
\caption{Installing \ecv and streaming from a camera with Pixi~\cite{pixi}.}
\label{lst:eventcv_stream}
\end{listing}

Beyond camera streaming, \ecv currently supports nine file formats read through the same function call. These file formats are HDF5, ROS bags of \code{dvs\_msgs}, iniVation AEDAT~2.0 and~4.0, Prophesee DAT and EVT2/EVT3 RAW, NumPy arrays, and delimited text files. There are two ways to read these files: \code{ecv.load} returns an \code{EventStream}, which loads all events into memory, while \code{ecv.open} creates an \code{EventReader}, which holds a plan for producing them slice by slice from a file, the more common method given the multi-GB size of event streams. Both support the same downstream operations, and \ecv can write every format it reads. We provide the \code{EventSink} class which enables live capture of an event stream while processing it simultaneously (Sec.~\ref{subsec:capture}).

\ecv is installable for Linux x86-64 and aarch64, macOS on Apple silicon, and Windows x86-64, with install solutions provided by PyPI\footnote{\url{https://pypi.org/project/eventcv/}} and conda-forge\footnote{\url{https://anaconda.org/channels/conda-forge/packages/eventcv/overview}}. aarch64 support matters for robotics, where many platforms use NVIDIA Jetson or Raspberry Pi. Providing binary packages rather than requiring source builds makes event cameras practical for roboticists.

\subsection{Capture and deployment for real-time robotics}
\label{subsec:capture}
Capture and decoding each run on a dedicated background thread. Inference loops collect windows of events and decode them for the forward pass. For robotics, PyTorch models are commonly exported to ONNX~\cite{onnxruntime} and run on compute platforms such as the Jetson, where runtime efficiency matters. Streaming from an event camera and running the events through an ONNX model is shown in Code snippet~\ref{lst:onnx}.

\begin{listing}[t]
\begin{minted}[frame=lines, breaklines, breakanywhere]{python}
model  = ecv.Model("detector.onnx") # or PyTorch module
camera = ecv.stream(dt_ms=5, repr="count")
while running:
    prediction = model(camera.read())
\end{minted}
\caption{ONNX model inference with \ecv.}
\label{lst:onnx}
\end{listing}

This function takes the last 5\,ms of events, builds an event-count representation, and runs a detector model specified in the \code{detector.onnx} file. Dense event streams are converted internally to NumPy arrays~\cite{Harris2020} such that PyTorch~\cite{Paszke2019} or ONNX can consume them without a memory-intensive copy. A model exported to ONNX for deployment loads events from \ecv directly, with batch axes inserted and integer representations cast as needed. For offline analysis, recordings can also be made alongside processing. Adding the \code{record="session.h5"} argument to the stream function writes the raw events from the capture thread while the loop works on model inference, resulting in deployments that also generate a replayable dataset at little additional cost.

\subsection{ROS~2 integration}
\label{subsec:ros2}
A library is usually connected to the rest of the robotics stack through ROS~2. As \ecv is written in Rust, we directly interface \ecv's core Rust library with ROS~2, for which there are two established options. \code{rclrs} is the official ROS~2 Rust client library\footnote{\url{https://github.com/ros2-rust/ros2_rust}}. \code{r2r}\footnote{\url{https://github.com/sequenceplanner/r2r}} provides simple Rust bindings for ROS~2. \ecv uses \emph{hiroz}~\cite{hiroz}, a Zenoh-native~\cite{zenoh} ROS~2 stack written in pure Rust. We use hiroz because \code{rclrs} and \code{r2r} both link the ROS C libraries, which requires a ROS installation at build time and prevents shipping the integration in a wheel. Zenoh on the wire means interoperating requires the modern \code{rmw\_zenoh\_cpp} middleware. A standard Data Distribution Service (DDS) deployment needs the \code{zenoh-bridge-ros2dds} in between.

Messages from an event camera are represented using standardized ROS~2 messages~\cite{rosevent}. We provide four nodes: a publisher of event packets, a subscriber that hands back the same \code{EventStream} the rest of the library takes, an image publisher that republishes a representation as a \code{sensor\_msgs/msg/Image}, and an inference node that runs a model and publishes its output as a \code{std\_msgs/msg/Float32MultiArray}. Replay and recording drive the publisher from a file and the subscriber to one. A voxel grid reaches the RViz2 visualizer without a conversion node.

\subsection{Inspecting event streams}
\label{subsec:gui}
Inspecting an event stream during playback is critical for evaluating camera settings, capture quality, and downstream usability. jAER~\cite{Delbruck2008}, Tonic~\cite{Lenz2021}, and Faery~\cite{faery} provide comprehensive visualization capabilities, with the ability to run filtering and processing directly within the visualization window.

\begin{figure}[!t]
    \centering
    \includegraphics[width=\columnwidth]{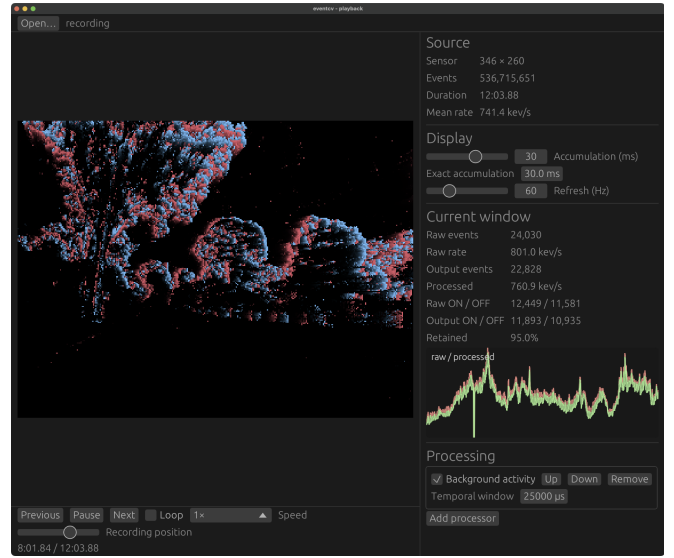}
    \caption{Graphical interface for viewing and live filtering event streams using \code{ecv.play}.}
    \label{fig:gui}
\end{figure}

\ecv provides a graphical user interface (GUI) to both replay recordings and view a live camera. An example is shown in Fig.~\ref{fig:gui}: there, the left panel accumulates events over a sliding window whose width and refresh rate are adjustable during playback. The right panel reports stream statistics and the number of events retained after background-activity filtering, over a rolling plot of the raw and processed rates. Filters can be applied during playback, giving immediate visual feedback.

\section{Representations, Filters, and Transforms}
\label{sec:reps}

\subsection{Representations}
\label{subsec:reps}
An event camera produces a sparse, ordered stream $\mathcal{E}$ of event tuples $e = (x, y, t, p)$, where $(x, y)$ are pixel coordinates, $t$ is a timestamp in microseconds, and $p \in \{-1, +1\}$ is the sign of the log-intensity change~\cite{Gallego2022}. For most computer vision applications, events are collated over a set time window $\Delta t$ or a maximum event count $N_{\max}$, compressing streams into dense arrays. \ecv implements the representations in common use, such as polarity histograms, voxel grids, and time surfaces. Recent representations developed for deep learning methods such as multi-channel time surfaces (MCTS)~\cite{Burkhardt2025} and tencode~\cite{Huang2023} are also implemented. Representations can be defined at the time of loading an event stream, or called explicitly on individual slices.

\begin{figure}[!t]
    \centering
    \includegraphics[width=\columnwidth]{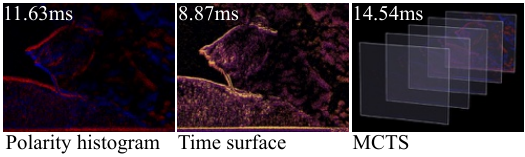}
    \caption{Three examples of the representations \ecv builds, from an N-ImageNet~\cite{Kim2021} stream. The top-left corner of each panel gives the time to generate it from 106,295 events over a 300\,ms window on a Jetson Orin AGX.}
    \label{fig:representations}
\end{figure}

Defining the representation at load time is simplest when only one is needed, whereas calling it on individual slices allows several to be mixed. Fig.~\ref{fig:representations} shows a selection of the representations \ecv builds and the time taken to generate them from an example event stream with 106,295 events captured over 300\,ms.

\subsection{Filters and geometric transforms}
\label{subsec:filters}
Event streams are inherently noisy. \ecv provides three common filters and a range of geometric transforms.

A \emph{hot pixel} filter removes all events at pixels whose count exceeds $\mu + n\sigma$ over the active pixels, where $\mu$ is the mean, $\sigma$ is the standard deviation, and $n$ is the threshold in standard deviations ($n = 3$ by default). The mask can be estimated once across a whole recording and then applied to any slice.
Hot pixel coordinates can also be supplied by providing a file of pixel coordinates for individual cameras. On a 3.9\,GB test recording from a DAVIS346, the filter identified 22 hot pixels accounting for 15.4\,\% of all events in the file. Applying the resulting mask cut a representative 33\,ms slice from 26{,}677 to 24{,}584 events.

A \emph{background activity} filter~\cite{Liu2015Spatiotemporal} removes isolated events by requiring recent activity in neighboring pixels. In \ecv, this neighborhood can be defined using either the eight surrounding pixels or only the four edge-adjacent pixels. A \emph{refractory} filter instead suppresses events that occur within a user-specified dead time of the previous event at the same pixel. In \ecv, the dead time is given in microseconds and can be measured from either the last emitted event, matching the behavior of the hardware, or from every arriving event, as Tonic~\cite{Lenz2021} and evlib~\cite{evlib} do.

Filters compose and stay lazy on a reader, so \code{ecv.open("events.h5", hot\_pixel\_filter=True)\allowbreak.background\_activity\_filter(1000)}, where the argument is a 1\,ms correlation window, reads nothing until a slice is requested.

In addition to these filters, \ecv provides access to basic geometric transforms including crop, flip, rotate, and translate, which are applied directly to the event stream rather than to the derived representations. The same transforms are also available as seeded augmentations for training: each draws its randomness from the seed and the slice index rather than a thread-local generator, so a \code{DataLoader} reproduces the same stream whatever the worker count or access order.

\section{Features, Motion, and Simulation}
\label{sec:features}

\subsection{Corner detection and feature learning}
\label{subsec:corners}
\ecv implements the eFAST~\cite{Mueggler2017} and Harris~\cite{Harris1988,Glover2022luvharris} corner detection algorithms. eFAST keeps an event when recent neighboring events form a continuous arc around it, indicating a corner-like feature. Harris keeps points where the accumulated event frame changes strongly in two directions, identified from large eigenvalues of the local second-moment matrix.

\ecv also implements FEAST~\cite{Afshar2020}, an unsupervised feature learner that acts similarly to a visual-word dictionary built with OpenCV\textquotesingle s \code{features2d}~\cite{Bradski2000}. For each event processed by FEAST, the local time-surface patch is normalized and matched to the nearest prototype within an adaptive threshold. A match moves the matched prototype towards the input and tightens its threshold, while a miss loosens all of the thresholds. The prototype bank converges on the recording's most common local patterns without any labels. These algorithms are callable with a small set of functions (Code snippet~\ref{lst:feast}).

\begin{listing}
\begin{minted}[frame=lines, breaklines, breakanywhere]{python}
stream = ecv.open("events.h5")
corners = stream.efast()
feast = ecv.FEAST(n_features=25, patch=11, tau_ms=30.0)
feast.fit(stream, epochs=3)
\end{minted}
\caption{eFAST corner detection and FEAST feature learning in \ecv.}
\label{lst:feast}
\end{listing}

\begin{figure}[!t]
    \centering
    \includegraphics[width=\columnwidth]{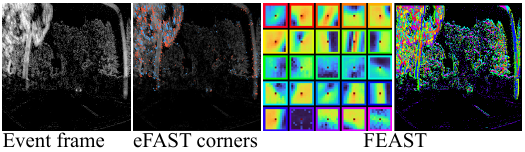}
    \vspace*{-0.3cm}
    \caption{Corner and feature detection on an event slice. Left to right: raw event frame, eFAST~\cite{Mueggler2017} corners, the learned FEAST~\cite{Afshar2020} prototype bank, and FEAST feature assignments.}
    \label{fig:features}
\end{figure}

Fig.~\ref{fig:features} demonstrates an example of an event frame with eFAST corner detections and learned FEAST features.

\subsection{Motion estimation and tracking}
\label{subsec:motion}
\begin{figure}[!t]
    \centering
    \includegraphics[width=\columnwidth]{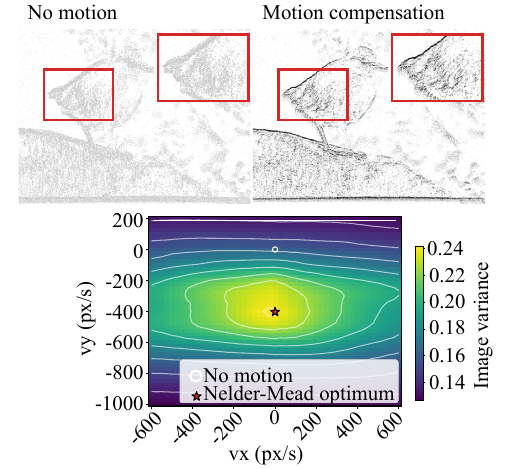}
    \vspace*{-0.5cm}
    \caption{Contrast-maximization motion compensation in \ecv. Top: an event slice before and after compensation, boxed regions magnified. Bottom: image variance over the translational velocity search, marking the uncompensated point (circle) and the Nelder-Mead optimum (star).}
    \label{fig:cmax}
\end{figure}

Contrast maximization recovers motion by searching for the warp that produces the sharpest Image of Warped Events (IWE)~\cite{Gallego2018} (Fig.~\ref{fig:cmax}). \ecv implements translational and rotational warp models and the three standard objectives: the IWE variance~\cite{Gallego2017}, the sum of squares, and the sum of exponentials~\cite{Stoffregen2019}. The warp parameters are found by Nelder-Mead simplex search.

\subsection{Simulation and reconstruction}
\label{subsec:sim}
Synthetic events provide a practical way to explore event data without a physical camera. \ecv can convert standard video files to event streams using a sensor model following video-to-events (v2e)~\cite{Hu2021}, with per-pixel threshold mismatch, an intensity-dependent photoreceptor low-pass filter, and leak and shot noise. A simulation is a single call, \code{ecv.simulate("video.mp4")}, requiring no model download or external repository.

\section{Evaluation}
\label{sec:evaluation}

This section tests three claims: that \ecv builds representations faster than existing libraries, that it decodes every format it supports at competitive throughput (both Sec.~\ref{subsec:reprthroughput}), and that it scales to modern sensor event rates without exhausting memory (Sec.~\ref{subsec:scaling}).

\subsection{Experimental setup}
\label{subsec:setup}
All measurements were run on a Jetson Orin AGX, an embedded platform representative of robotics deployments. The exception is the Event-LAB case study (Sec.~\ref{subsec:eventlab}), which ran entirely on an x86 desktop with an Intel i7-9700K and an NVIDIA RTX 2080. The inputs are a 3.0\,s, 1280$\times$720 EVT3 recording of 116.3\,M events (341\,MB on disk) and a 651\,s, 346$\times$260 HDF5 recording of 707.6\,M events (5.45\,GB on disk).

Times are the median of nine runs after two warm-ups, with the file in the page cache. Baselines are Tonic~1.6.0~\cite{Lenz2021}, event\_utils~\cite{Stoffregen2019}, dv-processing~2.0.4~\cite{dvprocessing}, evlib~0.13.2~\cite{evlib}, and Expelliarmus~1.1.7~\cite{expelliarmus}; each experiment uses those baselines that implement the operation under test. All baselines were run on the same machine, each under its own interpreter on identical input. All timings are CPU. \ecv's GPU path is limited to handing representations to PyTorch or ONNX Runtime on the device, and the \textit{Rust + GPU} entry in Table~\ref{tab:comparison} refers to that path rather than to GPU-accelerated event processing.

\subsection{Representation and decoding throughput}
\label{subsec:reprthroughput}
Event streams are often converted into frame-based representations for downstream processing and compatibility with models. When streaming from event cameras for real-time inference at or above 30\,Hz, efficient decoding and frame building are critical for edge deployment. We compare \ecv to the baseline methods in Table~\ref{tab:repr-timing}, measuring the time to build each representation from a 1.22\,M-event slice of the EVT3 recording and the rate at which each format is decoded. Over the representations more than one library provides, \ecv is 3.4$\times$ faster than evlib in geometric mean and 4.2$\times$ faster than Tonic, with per-representation speed-ups against the next-fastest library ranging from 2.8$\times$ to 3.7$\times$. The remaining representations, which neither Tonic nor evlib provide, build in 8.56--74.85\,ms.

    \begin{table}[!t]
      \footnotesize
      \centering
      \caption{Representation building time (ms, lower is better) from a 1.22\,M-event slice of the EVT3 recording. Speed-up is \ecv{} against the next-fastest library. ch.: channels.}
      \label{tab:repr-timing}
      \begin{tabular*}{\columnwidth}{@{\extracolsep{\fill}}lrrrr@{}}
        \toprule
        Representation & \ecv & Tonic~\cite{Lenz2021} & evlib~\cite{evlib} & Speed-up \\
        \midrule
        Polarity frame (2 ch.) & 23.78 & 136.36 &  67.22 & 2.8$\times$ \\
        Voxel grid (5 bins)    & 26.50 & 124.44 &  97.23 & 3.7$\times$ \\
        Time surface           & 21.33 &  59.16 &  80.76 & 2.8$\times$ \\
        \bottomrule
      \end{tabular*}
    \end{table}

Offline playback from recorded files is a common way of running event-processing pipelines, and decoding cost varies with the file format. We next compared the time taken to decode various file formats against every reader on the platform able to open each one (Table~\ref{tab:decode}). \ecv is the fastest reader on all three formats: EVT3 at 53.8\,Mev/s against Expelliarmus's 33.5\,Mev/s, AEDAT~4.0 at 85.4\,Mev/s against dv-processing's 77.0\,Mev/s on a file written by dv-processing's own LZ4 writer, and HDF5 at 16.3\,Mev/s against evlib's 5.1\,Mev/s. 

These results, however, only consider single-threaded operations. \ecv has the capacity to scale across cores for AEDAT~4.0 decoding, reaching 631.6\,Mev/s processing on twelve cores. evlib~\cite{evlib}, whose core is also Rust, decodes EVT3 an order of magnitude slower, and while it is the only baseline besides \ecv reading more than one container format, it does not support the AEDAT~4.0 file.

    \begin{table}[!t]
      \footnotesize
      \centering
      \caption{Decode throughput (Mev/s, higher is better), single-threaded, for every reader able to open each format. A dash marks a format the reader cannot open. For \ecv{}, the speed-up over the next-fastest reader for each format is shown in parentheses.}
      \label{tab:decode}
      \begin{tabular*}{\columnwidth}{@{\extracolsep{\fill}}lrrr@{}}
        \toprule
        Library & EVT3 raw & AEDAT 4.0 & HDF5 \\
                & \scriptsize 116.3\,Mev & \scriptsize 10.0\,Mev & \scriptsize 183.9\,Mev \\
        \midrule
        \ecv
          & \begin{tabular}[c]{@{}r@{}}\textbf{53.8}\\[-2pt]\scriptsize(1.6$\times$)\end{tabular}
          & \begin{tabular}[c]{@{}r@{}}\textbf{85.4}\\[-2pt]\scriptsize(1.1$\times$)\end{tabular}
          & \begin{tabular}[c]{@{}r@{}}\textbf{16.3}\\[-2pt]\scriptsize(3.2$\times$)\end{tabular} \\
        Expelliarmus~\cite{expelliarmus}
          & 33.5 & --- & --- \\
        dv-processing~\cite{dvprocessing}
          & --- & 77.0 & --- \\
        evlib~\cite{evlib}
          & 5.4 & --- & 5.1 \\
        \bottomrule
      \end{tabular*}
      \vspace{2pt}
    \end{table}

\subsection{Scaling and memory}
\label{subsec:scaling}
\begin{figure}[!t]
    \centering
    \includegraphics[width=\linewidth]{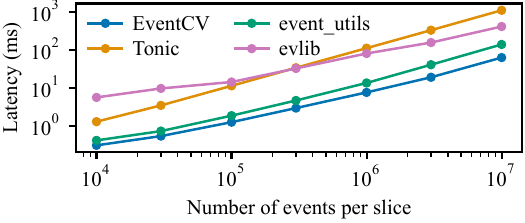}
    \caption{Latency of event counting against slice size, comparing \ecv with Tonic~\cite{Lenz2021}, event\_utils~\cite{Stoffregen2019}, and evlib~\cite{evlib} on a Jetson Orin AGX. Both axes are logarithmic.}
    \label{fig:bench}
\end{figure}

Different event cameras produce varying amounts of events per second, with larger sensors such as the Sony IMX636 capable of over 20\,Mev/s. To compare how different libraries scale to increasing event rates, we measured the latency to perform event counts across increasing event slice sizes (Fig.~\ref{fig:bench}). All libraries scale similarly with increasing event counts, and \ecv is the fastest at every slice size tested.

Every decode and representation timing above is measured through the same lazy reader. Its value shows on long recordings: on the 707.6\,M-event HDF5 recording, reading eagerly peaks at 8.87\,GB of resident memory, 13.5 bytes per event, because the packed on-disk record expands into four in-memory columns. Opening the same file lazily takes 0.11\,s, exposes it as 19{,}737 windows of 33\,ms, and peaks at 273\,MB while serving 200 windows drawn at random: a 33$\times$ reduction, with a random window retrieved in 8.6\,ms (16.0\,ms at the 99th percentile).

\section{Case Studies}
\label{sec:cases}

In this section, we present three case studies using \ecv. All evaluations were performed on a Jetson Orin AGX, except the Event-LAB study (Sec.~\ref{subsec:eventlab}), which ran entirely on the x86 desktop described in Sec.~\ref{subsec:setup}: both the \ecv CPU path and the CUDA path.

\subsection{Edge inference with an event-based YOLO detector}
\label{subsec:evyolo}
\begin{figure}[!t]
    \centering
    \includegraphics[width=0.8\linewidth]{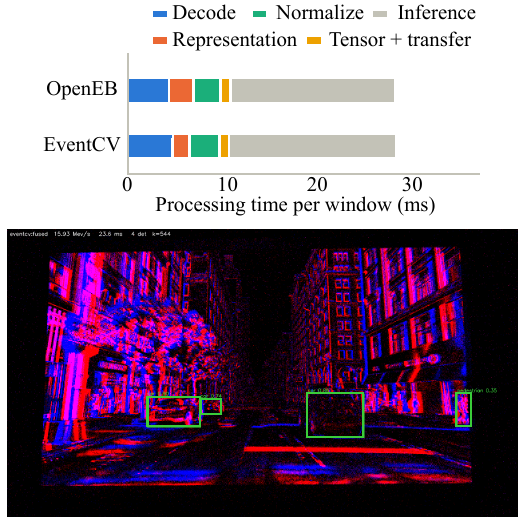}
    \vspace{-5pt}
    \caption{The ev-ultralytics YOLO26~\cite{yolov8_ultralytics,ev_ultralytics_prophesee} pipeline on an EVK4 recording, run on a Jetson Orin AGX. Top: per-window processing time by stage, OpenEB against \ecv. Bottom: example detections from the \ecv pipeline.}
    \label{fig:evyolo}
\end{figure}

ev-ultralytics~\cite{yolov8_ultralytics,ev_ultralytics_prophesee} is Prophesee's adaptation of YOLO26 to event cameras for object detection. Briefly, a polarity histogram of events is accumulated over 50\,ms, quantized to \code{uint8}, converted to a tensor format, and passed to the object detector's inference model. To measure the latency of the official open-source SDK from conversion of an event stream to model prediction, we implemented the same pipeline and replaced the OpenEB backend with \ecv. Fig.~\ref{fig:evyolo} compares the per-window processing time by stage.

Both pipelines reach the network tensor in about 10.5\,ms per window and complete in about 28\,ms, of which roughly 17.5\,ms is the shared YOLO26 inference: \ecv matches the OpenEB \code{EventsIterator} pipeline stage for stage rather than beating it. The difference is in what it costs to build and deploy. The \ecv pipeline is seven lines of user code and a single library call, compared to the OpenEB pipeline that requires a custom-written accumulator. The installation pathway differs sharply. \ecv is a \code{pip/conda/pixi install eventcv}. OpenEB's Python bindings come from a third-party apt repository including 44 packages tied to the system-level Python without prebuilt ARM packages, limiting Jetson deployment. We note that OpenEB is Prophesee's archived open-source SDK that has been replaced with the newer proprietary Metavision.

\subsection{Live keypoint detection with SuperEvent}
\label{subsec:motionest}
\begin{figure}[!t]
    \centering
    \includegraphics[width=0.8\linewidth]{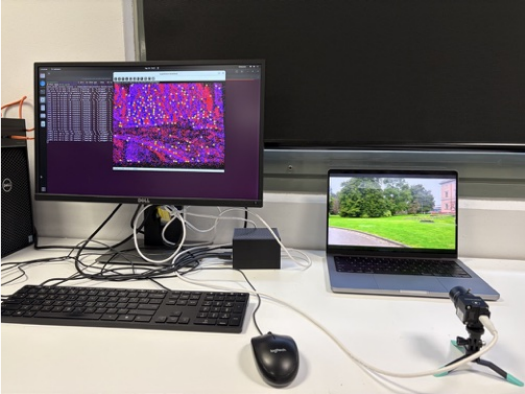}
    \vspace{-5pt}
    \caption{Capture rig for the SuperEvent~\cite{Burkhardt2025} case study. A Prophesee EVK4 views an outdoor walking sequence replayed on a laptop screen. Events stream through \ecv into SuperEvent on a Jetson Orin AGX.}
    \label{fig:superevent}
\end{figure}

\ecv can run ONNX-compiled models directly on a live event camera stream. We evaluated the ability to run live keypoint detection using SuperEvent~\cite{Burkhardt2025} streaming directly from a Prophesee EVK4 (Fig.~\ref{fig:superevent}). Each inference used a 50\,ms accumulation window and the pre-trained SuperEvent weights. A recording of an outdoor walking environment was captured on a laptop screen, yielding up to 170 keypoints per window.

End-to-end processing took 44.9--71.7\,ms per inference, corresponding to 14--22\,Hz. This covers building the multi-channel time surface, resizing it to the input dimensions, and running the model. \ecv runs any exported ONNX model directly on a camera stream (Code snippet~\ref{lst:onnx}).

\subsection{Event-LAB localization pipeline}
\label{subsec:eventlab}
\begin{figure}[!t]
    \centering
    \includegraphics[width=\linewidth]{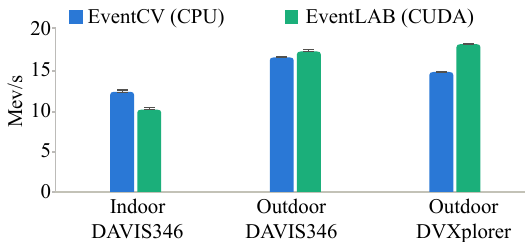}
    \vspace{-15pt}
    \caption{Comparison of \ecv deployed on CPU with custom CUDA data formatting in Event-LAB~\cite{Hines2025eventlab}. Throughput in Mev/s, higher is better.}
    \label{fig:eventlab}
\end{figure}

Event-LAB~\cite{Hines2025eventlab} is a framework for standardized evaluation of event-based localization. To implement it, more than 2,700 lines of custom code were developed to standardize file formats to \code{.h5}, build event representations for the various localization methods, and run event-to-video generation~\cite{Rebecq2021}. A large part of the dataset formatting code consisted of custom implementations of CUDA-accelerated event representation building to improve runtimes. Fig.~\ref{fig:eventlab} compares the frame generation runtime for \ecv against the custom CUDA implementation. \ecv on the CPU delivers throughput comparable to the custom CUDA implementation across all three recordings, so a GPU is not required for this stage of the pipeline. Re-implementing Event-LAB with \ecv removes this overhead.

Another difference is that the original Event-LAB dataset formatting code treated representation generation sequentially. As a result, starting a localization run at an offset forced the code to iterate through the file manually until it reached a user-specified timestamp. With \ecv, this is now resolved by using the lazy slicer \code{ecv.open()} to identify the exact slice to begin from.

\section{Discussion and Conclusions}
\label{sec:limitations}
In this work, we presented \ecv, an open-source, extensible, and efficient computer vision package for working with event cameras. Through three case studies, we showed that \ecv can be used for a variety of robotic vision applications without custom decoding or representation code, and without relying on multiple external packages.

\ecv was designed specifically for computer vision tasks on event data, covering camera streaming, file I/O, filtering, representations, features, motion estimation, and inference in one package. It is tailored to lower the barrier to working with event cameras, inspired by open-source tools for conventional images such as OpenCV~\cite{Bradski2000} and Pillow~\cite{pillow}.

Many other open-source event camera libraries exist (Table~\ref{tab:comparison}). \ecv provides a more comprehensive and efficient system for use in modern machine learning and robotics workflows. \ecv is intended to support open collaboration through the integration of existing workflows, community outreach, and education. This includes high-quality tutorial content and proposals for workshop tutorials at major robotics and computer vision conferences.

\ecv has limitations. GPU acceleration is currently limited to handing representations to PyTorch or ONNX Runtime, so event processing itself remains CPU-bound. The ROS~2 integration goes over Zenoh, so a standard DDS deployment needs a bridge in between. Feature learning and motion estimation are re-implementations of established methods, and we have not evaluated them against their reference implementations.

\ecv is an open-source and extensible Python package for event-based computer vision applications with an efficient Rust core and thin Python API. We have shown its utility in streaming from event cameras, processing recorded files, and simplifying existing workflows.

\let\oldthebibliography\thebibliography
\renewcommand{\thebibliography}[1]{%
  \oldthebibliography{#1}%
  \setlength{\itemsep}{0pt}%
  \setlength{\parskip}{0pt}%
}
\bibliography{Bibliography}

\begin{thebibliography}{10}
\providecommand{\url}[1]{#1}
\csname url@samestyle\endcsname
\providecommand{\newblock}{\relax}
\providecommand{\bibinfo}[2]{#2}
\providecommand{\BIBentrySTDinterwordspacing}{\spaceskip=0pt\relax}
\providecommand{\BIBentryALTinterwordstretchfactor}{4}
\providecommand{\BIBentryALTinterwordspacing}{\spaceskip=\fontdimen2\font plus
\BIBentryALTinterwordstretchfactor\fontdimen3\font minus \fontdimen4\font\relax}
\providecommand{\BIBforeignlanguage}[2]{{%
\expandafter\ifx\csname l@#1\endcsname\relax
\typeout{** WARNING: IEEEtran.bst: No hyphenation pattern has been}%
\typeout{** loaded for the language `#1'. Using the pattern for}%
\typeout{** the default language instead.}%
\else
\language=\csname l@#1\endcsname
\fi
#2}}
\providecommand{\BIBdecl}{\relax}
\BIBdecl

\bibitem{Bradski2000}
G.~Bradski, ``The {OpenCV} library,'' \emph{Dr. Dobb's J. Softw. Tools}, vol.~25, no.~11, pp. 120, 122--125, 2000.

\bibitem{Harris2020}
C.~R. Harris \emph{et~al.}, ``Array programming with {NumPy},'' \emph{Nature}, vol. 585, pp. 357--362, 2020.

\bibitem{Paszke2019}
A.~Paszke \emph{et~al.}, ``{PyTorch}: An imperative style, high-performance deep learning library,'' in \emph{Adv. Neural Inf. Process. Syst. (NeurIPS)}, 2019.

\bibitem{Corke2005}
P.~I. Corke, ``The machine vision toolbox,'' \emph{IEEE Robotics \& Automation Magazine}, vol.~12, no.~4, pp. 16--25, 2005.

\bibitem{Lichtsteiner2008}
P.~Lichtsteiner, C.~Posch, and T.~Delbruck, ``A 128$\times$128 120 {dB} 15 $\mu$s latency asynchronous temporal contrast vision sensor,'' \emph{IEEE J. Solid-State Circuits}, vol.~43, no.~2, pp. 566--576, 2008.

\bibitem{Brandli2014}
C.~Brandli \emph{et~al.}, ``A 240$\times$180 130 {dB} 3 $\mu$s latency global shutter spatiotemporal vision sensor,'' \emph{IEEE J. Solid-State Circuits}, vol.~49, no.~10, pp. 2333--2341, 2014.

\bibitem{Gallego2022}
G.~Gallego \emph{et~al.}, ``Event-based vision: A survey,'' \emph{IEEE Trans. Pattern Anal. Mach. Intell.}, vol.~44, no.~1, pp. 154--180, 2022.

\bibitem{dvprocessing}
{iniVation AG}, ``{dv-processing}: Generic processing algorithms for event cameras,'' \url{gitlab.com/inivation/dv/dv-processing}, 2023.

\bibitem{metavision}
{Prophesee}, ``{Metavision SDK},'' \url{docs.prophesee.ai}, 2024.

\bibitem{aedat}
{International Centre for Neuromorphic Systems} and A.~Marcireau, ``aedat: A fast {AEDAT4} decoder with an underlying {Rust} implementation,'' \url{github.com/neuromorphicsystems/aedat}, 2025.

\bibitem{Pedersen2023}
J.~E. Pedersen and J.~Conradt, ``Aestream: Accelerated event-based processing with coroutines,'' in \emph{Neuro-Inspired Computational Elements Conf. (NICE)}, 2023, p. 86–91.

\bibitem{evlib}
T.~Allam~Jr., ``{evlib}: Event camera data processing library,'' \url{github.com/tallamjr/evlib}, 2026.

\bibitem{Stoffregen2019}
T.~Stoffregen and L.~Kleeman, ``Event cameras, contrast maximization and reward functions: An analysis,'' in \emph{IEEE/CVF Conf. Comput. Vis. Pattern Recognit. (CVPR)}, 2019, pp. 12\,292--12\,300.

\bibitem{expelliarmus}
F.~Ottati and G.~Lenz, ``{expelliarmus}: A {Python} package for decoding prophesee {RAW} and {DAT} event files,'' \url{github.com/open-neuromorphic/expelliarmus}, 2023, archived September 2024.

\bibitem{faery}
A.~Marcireau \emph{et~al.}, ``{Faery}: A stream processing library for neuromorphic event-based data,'' \url{github.com/aestream/faery}, 2025.

\bibitem{Delbruck2008}
T.~Delbruck, ``Frame-free dynamic digital vision,'' in \emph{Int. Symp. Secure-Life Electron.}, 2008, pp. 21--26.

\bibitem{Lenz2021}
\BIBentryALTinterwordspacing
G.~Lenz \emph{et~al.}, ``Tonic: event-based datasets and transformations.'' Jul. 2021, {Documentation available under https://tonic.readthedocs.io}. [Online]. Available: \url{https://doi.org/10.5281/zenodo.5079802}
\BIBentrySTDinterwordspacing

\bibitem{Rebecq2018esim}
H.~Rebecq, D.~Gehrig, and D.~Scaramuzza, ``{ESIM}: an open event camera simulator,'' in \emph{Conf. Robot Learning (CoRL)}, ser. Proc. Mach. Learn. Res., vol.~87, 2018, pp. 969--982.

\bibitem{neuromorphicdrivers}
{International Centre for Neuromorphic Systems} and A.~Marcireau, ``{Neuromorphic Drivers}: {Python} and {Rust} libraries to interact with event cameras in real-time,'' \url{github.com/neuromorphicsystems/neuromorphic-drivers}, 2023.

\bibitem{Glover2018yarp}
A.~Glover \emph{et~al.}, ``The event-driven software library for {YARP}---with algorithms and {iCub} applications,'' \emph{Front. Robot. AI}, vol.~4, 2018.

\bibitem{Mueggler2017ros}
E.~Mueggler \emph{et~al.}, ``\texttt{rpg\_dvs\_ros}: {ROS} driver packages for {DVS}/{DAVIS} event cameras,'' \url{github.com/uzh-rpg/rpg_dvs_ros}, 2020.

\bibitem{rosevent}
B.~Pfrommer, ``{ROS}~2 event camera stack: Messages, codecs and drivers for {Prophesee}, {CenturyArks} and {iniVation} sensors,'' \url{github.com/ros-event-camera}, 2026.

\bibitem{Kim2021}
J.~Kim \emph{et~al.}, ``{N-ImageNet}: Towards robust, fine-grained object recognition with event cameras,'' in \emph{IEEE/CVF Int. Conf. Comput. Vis. (ICCV)}, 2021, pp. 2146--2156.

\bibitem{Mueggler2017}
E.~Mueggler, C.~Bartolozzi, and D.~Scaramuzza, ``Fast event-based corner detection,'' in \emph{British Mach. Vis. Conf. (BMVC)}, 2017, pp. 33.1--33.11.

\bibitem{Alzugaray2018}
I.~Alzugaray and M.~Chli, ``Asynchronous corner detection and tracking for event cameras in real time,'' \emph{IEEE Robot. Autom. Lett.}, vol.~3, no.~4, pp. 3177--3184, 2018.

\bibitem{Afshar2020}
S.~Afshar \emph{et~al.}, ``Event-based feature extraction using adaptive selection thresholds,'' \emph{Sensors}, vol.~20, no.~6, p. 1600, 2020.

\bibitem{Gallego2017}
G.~Gallego and D.~Scaramuzza, ``Accurate angular velocity estimation with an event camera,'' \emph{IEEE Robot. Autom. Lett.}, vol.~2, no.~2, pp. 632--639, 2017.

\bibitem{Gallego2018}
G.~Gallego, H.~Rebecq, and D.~Scaramuzza, ``A unifying contrast maximization framework for event cameras, with applications to motion, depth, and optical flow estimation,'' in \emph{IEEE/CVF Conf. Comput. Vis. Pattern Recognit. (CVPR)}, 2018, pp. 3867--3876.

\bibitem{Stoffregen2021}
T.~Stoffregen, ``Motion estimation by focus optimisation: Optic flow and motion segmentation with event cameras,'' Ph.D. dissertation, Department of Electrical and Computer Systems Engineering, Monash University, 2021.

\bibitem{Zhu2019}
A.~Z. Zhu \emph{et~al.}, ``Unsupervised event-based learning of optical flow, depth, and egomotion,'' in \emph{IEEE/CVF Conf. Comput. Vis. Pattern Recognit. (CVPR) Workshops}, 2019, pp. 1694--1694.

\bibitem{pixi}
\BIBentryALTinterwordspacing
T.~Fischer \emph{et~al.}, ``Pixi: Unified software development and distribution for robotics and ai,'' 2025. [Online]. Available: \url{https://arxiv.org/abs/2511.04827}
\BIBentrySTDinterwordspacing

\bibitem{onnxruntime}
O.~R. developers, ``{ONNX} {R}untime,'' \url{https://onnxruntime.ai/}, 2021.

\bibitem{hiroz}
{ZettaScale Technology}, ``\emph{hiroz}: A {Zenoh}-native {ROS}~2 stack in pure {Rust},'' \url{github.com/ZettaScaleLabs/hiroz}, 2026.

\bibitem{zenoh}
A.~Corsaro \emph{et~al.}, ``Zenoh: Unifying communication, storage and computation from the cloud to the microcontroller,'' in \emph{Euromicro Conf.\ Digital System Design (DSD)}, 2023, pp. 422--428.

\bibitem{Burkhardt2025}
Y.~Burkhardt, S.~Schaefer, and S.~Leutenegger, ``{SuperEvent}: Cross-modal learning of event-based keypoint detection for {SLAM},'' in \emph{Proceedings of the IEEE/CVF International Conference on Computer Vision (ICCV)}, 2025, pp. 8918--8928.

\bibitem{Huang2023}
Z.~Huang \emph{et~al.}, ``{EventPoint}: Self-supervised interest point detection and description for event-based camera,'' in \emph{IEEE/CVF Winter Conf. Appl. Comput. Vis. (WACV)}, 2023, pp. 5396--5405.

\bibitem{Liu2015Spatiotemporal}
H.~Liu \emph{et~al.}, ``Design of a spatiotemporal correlation filter for event-based sensors,'' in \emph{2015 IEEE International Symposium on Circuits and Systems (ISCAS)}, 2015, pp. 722--725.

\bibitem{Harris1988}
C.~Harris and M.~Stephens, ``A combined corner and edge detector,'' in \emph{Proc. Alvey Vis. Conf.}, 1988, pp. 147--151.

\bibitem{Glover2022luvharris}
A.~Glover \emph{et~al.}, ``{luvHarris}: A practical corner detector for event-cameras,'' \emph{IEEE Trans. Pattern Anal. Mach. Intell.}, vol.~44, no.~12, pp. 10\,087--10\,098, 2022.

\bibitem{Hu2021}
Y.~Hu, S.-C. Liu, and T.~Delbruck, ``v2e: From video frames to realistic {DVS} events,'' in \emph{IEEE/CVF Conf. Comput. Vis. Pattern Recognit. (CVPR) Workshops}, 2021, pp. 1312--1321.

\bibitem{yolov8_ultralytics}
\BIBentryALTinterwordspacing
G.~Jocher, A.~Chaurasia, and J.~Qiu, ``Ultralytics {YOLO},'' 2025. [Online]. Available: \url{https://github.com/ultralytics/ultralytics}
\BIBentrySTDinterwordspacing

\bibitem{ev_ultralytics_prophesee}
{Prophesee}, ``Ev-ultralytics: Ultralytics yolo adapted for event-based object detection,'' \url{https://github.com/prophesee-ai/ev-ultralytics}, 2024, accessed: 2026-09-06.

\bibitem{Hines2025eventlab}
A.~D. Hines \emph{et~al.}, ``{Event-LAB}: Towards standardized evaluation of neuromorphic localization methods,'' in \emph{IEEE Int. Conf. Robot. Autom. (ICRA)}, 2026.

\bibitem{Rebecq2021}
H.~Rebecq \emph{et~al.}, ``High speed and high dynamic range video with an event camera,'' \emph{IEEE Trans. Pattern Anal. Mach. Intell.}, vol.~43, pp. 1964--1980, 2021.

\bibitem{pillow}
\BIBentryALTinterwordspacing
A.~Clark and contributors, ``Pillow (python imaging library).'' [Online]. Available: \url{https://python-pillow.org}
\BIBentrySTDinterwordspacing

\end{thebibliography}
\end{document}